\documentclass[lettersize,journal]{IEEEtran}
\usepackage{amsmath,amsfonts}
\usepackage{algorithmic}
\usepackage{algorithm}
\usepackage{array}
\usepackage[caption=false,font=normalsize,labelfont=sf,textfont=sf]{subfig}
\usepackage{textcomp}
\usepackage{stfloats}
\usepackage{url}
\usepackage{verbatim}
\usepackage{graphicx}
\usepackage{cite}
\usepackage{xcolor}         
\usepackage{multirow}
\usepackage{booktabs} 
\usepackage{bm}
\usepackage[colorlinks,citecolor=blue,urlcolor=blue,linkcolor=blue]{hyperref}

\begin{document}


\title{OKG-LLM: Aligning Ocean Knowledge Graph with Observation Data via LLMs for Global Sea Surface Temperature Prediction}

\author{IEEE Publication Technology,~\IEEEmembership{Staff,~IEEE,}
}
\author{Hanchen Yang, Jiaqi Wang, Jiannong Cao,~\IEEEmembership{Fellow,~IEEE}, Wengen Li, Jialun Zheng, \\ Yangning Li, Chunyu Miao,  Jihong Guan, Shuigeng Zhou, and Philip S. Yu,~\IEEEmembership{Fellow,~IEEE}.        
\thanks{Hanchen Yang is with the Department of Computer Science and Technology, Tongji University, Shanghai, China, and also the Department of Computing, The Hong Kong Polytechnic University, Hong Kong, China~(neoyang@tongji.edu.cn). Jiaqi Wang, Wengen Li, and Jihong Guan are with the Department of Computer Science and Technology, Tongji University, Shanghai, China~(wangjq, lwengen, jhguan@tongji.edu.cn). Jiannong Cao and Jialun Zheng are with the Department of Computing, The Hong Kong Polytechnic University, Hong Kong, China~(jiannong.cao, jialun.zheng@polyu.edu.hk). Yangning Li, Chunyu Miao, and Philip S. Yu are with the Department of Computing, University of Illinois Chicago, Chicago, United States~(liyangning98@gmail.com, cmiao8, psyu@uic.edu). Shuigeng Zhou is with the School of Computer Science, Fudan University, Shanghai, China~(sgzhou@fudan.edu.cn).}
\thanks{Hanchen Yang and Jiaqi Wang contributed equally to this work, Wengen Li is the corresponding author.}}

\markboth{Journal of \LaTeX\ Class Files,~Vol.~XX, No.~XX,~XX}%
{Shell \MakeLowercase{\textit{et al.}}: A Sample Article Using IEEEtran.cls for IEEE Journals}


\maketitle

\begin{abstract}
Sea surface temperature (SST) prediction is a critical task in ocean science, supporting various applications, such as weather forecasting, fisheries management, and storm tracking. While existing data-driven methods have demonstrated significant success, they often neglect to leverage the rich domain knowledge accumulated over the past decades,
limiting further advancements in prediction accuracy.
The recent emergence of large language models (LLMs) has highlighted the potential of integrating domain knowledge for downstream tasks. 
However, the application of LLMs to SST prediction remains underexplored, primarily due to the challenge of integrating ocean domain knowledge and numerical data.
To address this issue, we propose Ocean Knowledge Graph-enhanced LLM (OKG-LLM), a novel framework for global SST prediction. 
To the best of our knowledge, this work presents the first systematic effort to construct an Ocean Knowledge Graph (OKG) specifically designed to represent diverse ocean knowledge for SST prediction.
We then develop a graph embedding network to learn the comprehensive semantic and structural knowledge within the OKG, capturing both the unique characteristics of individual sea regions and the complex correlations between them.
Finally, we align and fuse the learned knowledge with fine-grained numerical SST data and leverage a pre-trained LLM to model SST patterns for accurate prediction.
Extensive experiments on the real-world dataset demonstrate that OKG-LLM consistently outperforms state-of-the-art methods, showcasing its effectiveness, robustness, and potential to advance SST prediction.
The codes are available in the online repository\footnote{https://github.com/Neoyanghc/OKGLLM}.

\end{abstract}

\begin{IEEEkeywords}
Sea surface temperature prediction, large language models, knowledge graph, knowledge and data alignment
\end{IEEEkeywords}

\section{Introduction}

\IEEEPARstart{S}{ea} surface temperature (SST) is a critical factor that significantly influences global climate change. Over the past decade, SST prediction has emerged as a vibrant interdisciplinary field bridging ocean science and computer science, and plays a pivotal role in enabling informed decision-making across various applications, such as weather forecasting, storm tracking, and fisheries management~\cite{bi2023accurate}.

As a typical time series prediction task, numerous data-driven methods~(e.g., CNN, RNN, Transformer) have been proposed for SST prediction~\cite{ma2024survey,yang2023spatial}. 
Despite their success, these data-driven methods often overlook essential domain knowledge developed over decades of ocean science research, limiting further advancements in prediction accuracy.
The well-established knowledge, such as climate zones, ocean currents, and ENSO (El Niño-Southern Oscillation) areas, is closely linked to SST patterns across various ocean regions~\cite{wang2022ocean, timmermann2018nino}.
For example, the Peru cold current, which connects the Eastern Pacific cold tongue, provides valuable insights into SST variations in the equatorial Eastern Pacific region~\cite{kessler2006circulation}. 
Therefore, the failure to leverage informative and semantic oceanic knowledge limits the potential of current data-driven approaches to further enhance global SST prediction.

Recently, large language models (LLMs)~(e.g., GPT~\cite{radford2018improving}, Llama~\cite{touvron2023llama}, and Deepseek~\cite{liu2024deepseek, liang2022mixed}) have rapidly evolved in academia and industry and exhibit exceptional capabilities in understanding and reasoning across diverse domains.
This success has spurred initial efforts to adapt LLMs for time-series prediction~\cite{zhou2023one, liu2024unitime, jin2023time, jin2024position}. For example, GPT4TS~\cite{zhou2023one} utilizes GPT-2 for feature embedding, and outperforms traditional methods.
Despite these advancements, current methodologies primarily leverage LLMs as powerful but generic feature extractors, rather than as reasoners equipped with domain-specific insights~\cite{ma2024survey}. This limited engagement with domain-specific knowledge signifies a critical yet underexplored direction, particularly in the SST prediction task, which relies heavily on rich ocean knowledge. 
Therefore, effectively integrating ocean knowledge with data-driven models represents a promising, albeit challenging, path toward advancing SST prediction.
However, to this end, two major challenges must be addressed:

\textbf{Challenge 1: Limited representation of ocean
knowledge and regional correlations.} 
Existing SST prediction methods face significant challenges in capturing semantic ocean knowledge and complex inter-regional correlations. 
Some recent approaches~\cite{liu2024unitime,jin2023time, caotempo,jin2024position} attempt to leverage LLMs to incorporate domain knowledge by using task description texts as prompts. However, these methods remain insufficient. For example, UniTime~\cite{liu2024unitime} uses textual prompts describing the dataset (e.g., “The input prediction task is to predict global sea surface temperature collected every week") as inputs to LLMs, which fail to capture rich semantic knowledge. In the ocean domain, there is a wealth of knowledge highly relevant to SST changes, such as climate zones, ocean currents, and monsoons.
Limited input fails to comprehensively represent the nuanced and specialized ocean knowledge of a given region. 
In addition, existing SST prediction methods typically identify regional correlations based on numerical data analysis~\cite {yang2023higrn, peng2025cross, peng2023mustc}, and often simplify these correlations to distance-based metrics, overlooking multifaceted and physically grounded connections—such as those driven by ocean currents and atmospheric teleconnections—that are not explicitly present in the raw numerical data. Therefore, existing data-driven SST prediction models are often limited in capturing the high-level correlations, thereby restricting prediction accuracy.

\textbf{Challenge 2: Heterogeneity between ocean knowledge and fine-grained SST data.} 
Another fundamental challenge is the modality difference and granularity discrepancy between the knowledge provided to LLMs and the numerical SST data. Current LLM-based prediction methods~\cite{liu2024unitime,jin2023time,zhou2023one} typically employ a single, coarse-grained textual prompt for the entire dataset. This "one-size-fits-all" strategy is at odds with the ocean's heterogeneity, where distinct regions like the equatorial Pacific and the North Atlantic exhibit vastly different SST patterns.
This mismatch creates a fundamental challenge—applying macroscopic knowledge to predict diverse and region-specific SST data. Existing SST prediction models usually fail to align the general information with the specific, detailed characteristics of different regions. As a result, they struggle to capture critical local dynamics. Effectively achieving fine-grained alignment between ocean knowledge and regional SST data is thus a key challenge for SST prediction. 
To address these two challenges, we propose the Ocean Knowledge Graph-enhanced LLM (OKG-LLM), a novel framework for global SST prediction.
Specifically, we first construct a comprehensive Ocean Knowledge Graph (OKG) to systematically integrate and represent heterogeneous knowledge derived from various oceanographic studies. The OKG encompasses five critical types of entities that exhibit strong correlations with SST variability, including ocean currents, climatic zones, monsoon systems, geographic regions, and special oceanic areas (e.g., upwelling zones).
We then develop an OKG encoding network to extract region-specific oceanic features and model the complex interdependencies among distinct marine regions. Additionally, we introduce a fine-grained knowledge alignment module that establishes explicit mappings between the structured OKG representations and numerical SST data, enabling effective knowledge infusion into the LLM framework for accurate SST prediction.
The OKG-LLM offers a structured, domain-specific knowledge framework that systematically captures distinctive oceanic patterns and their dynamic interactions, ultimately enhancing SST prediction accuracy.


To summarize, the main contributions of this paper are:
\begin{itemize}
    \item \textbf{Development of the OKG-LLM framework}: We propose OKG-LLM, a novel model that integrates semantic and structural ocean knowledge with real-world observational data to enhance SST prediction using LLMs. To the best of our knowledge, this is the first attempt to unify domain-specific ocean knowledge with observational data for SST forecasting.
    \item \textbf{Construction of the Ocean Knowledge Graph (OKG)}: We develop the Ocean Knowledge Graph (OKG), the first fine-grained, open-source knowledge graph explicitly designed to support SST prediction in ocean science. As a pioneering effort, we anticipate that the insights and results derived from OKG will 
    benefit a range of data-driven approaches in the ocean science domain.
    \item \textbf{Comprehensive empirical validation}: Extensive experiments on real-world global SST datasets demonstrate that OKG-LLM achieves state-of-the-art performance across nine baselines, highlighting its superiority.
\end{itemize}


The remainder of this paper is organized as follows. Section 2 provides a comprehensive review of related work, and
Section 3 presents the technical details of OKG-LLM.
In Section 4, we conduct rigorous empirical evaluation on real-world SST datasets to validate the effectiveness of the proposed model. Finally, Section 5 concludes this work and suggests promising directions for future research.


\section{Related Work}
In this section, we review the literature on SST prediction, LLM for time series prediction, and domain knowledge enhanced prediction.

\textbf{Sea surface temperature prediction.}
Existing approaches for SST prediction can be categorized into two primary paradigms: numerical models~\cite{meng2023physical,yang2023spatial} and data-driven models~\cite{chen2023multi, 10726725, jin2023evaluating}. 
Physics-based numerical models, such as the ECMWF system~\cite{molteni1996ecmwf}, use oceanographic principles to model system dynamics but are often computationally expensive and constrained by strong assumptions. In contrast, data-driven models, particularly deep learning models, have gained prominence by leveraging historical data to extract complex temporal dependencies. Architectures such as Recurrent Neural Networks (RNNs) and Transformers are commonly employed for this purpose~\cite{10726725, 10974997, jin2022multivariate}. Advanced models like Informer~\cite{zhou2021informer}, Fedformer~\cite{zhou2022fedformer}, and TimeXer~\cite{wang2024timemixer} have enhanced the Transformer architecture specifically for long-term time series forecasting. Furthermore, other studies such as DLinear~\cite{zeng2023transformers} and FreTS~\cite{yi2023frequency} have demonstrated that simpler linear models can also achieve competitive performance, questioning the necessity of complex architectures for all forecasting tasks. Although these data-driven models have shown potential, they often overlook critical oceanographic knowledge, which limits their prediction accuracy. Instead, our method provides a more effective framework that first constructs the ocean knowledge graph and then fuses the domain knowledge with numerical data for enhancing the prediction performance.


\textbf{LLM for time series prediction.}
Inspired by the remarkable achievements of LLMs in broad domains like CV and NLP, leveraging LLM for time series prediction has garnered attention as a promising direction~\cite{10945394}. PromptCast~\cite{xue2022promptcast} and OFA~\cite{zhou2023one} are two early attempts to leverage LLMs for general time series analysis, where the former is purely based on prompting while the latter fine-tunes LLMs for downstream tasks. A recent work, Time-LLM~\cite{jin2023time}, reprograms time series and integrates natural language prompts to unleash the full potential of off-the-shelf LLMs. For embedding spatial dependencies, UrbanGPT\cite{li2024urbangpt} introduces the spatial instruction-tuning paradigm to add spatial information in the prompt and achieves promising performance in urban-related tasks. However, by treating LLMs as generic feature extractors, these methods overlook the domain-specific knowledge of time-series data. Moreover, the heterogeneity between ocean knowledge and fine-grained SST data is still underexplored. Our work introduces a fine-grained alignment method designed specifically to fuse these two data types, leading to more accurate predictions.

\textbf{Domain knowledge enhanced prediction.}
In the era of foundation models, the integration of domain-specific knowledge is essential for enhancing model performance on downstream tasks across numerous fields such as bioscience~\cite{fu2025foundation,ferruz2022protgpt2}, geoscience~\cite{deng2024k2}, and materials science~\cite{takeda2023foundation, wang2022leveraging}. 
Similar efforts are emerging in ocean science~\cite{10.1145/3654777.3676462,bi2024oceangpt}.
For example, OceanGPT~\cite{bi2024oceangpt} fine-tunes a pre-trained Large Language Model (LLM) with an extensive ocean domain corpus to achieve strong performance in ocean-related question answering and understanding. 
However, methods for integrating domain knowledge with time-series data to improve ocean predictions have not yet been developed.

Knowledge Graphs (KGs), as a powerful tool for representing domain knowledge, have been applied in many applications~\cite {xue2022knowledge,10387715}.
By representing entities and their interconnections in a structured manner, KGs provide deeper insights and enable more accurate predictions in various domains than traditional data-driven approaches~\cite{yao2024knowledge,peng2023knowledge}. 
For example, UUKG~\cite{ning2023uukg} presents a unified urban knowledge graph dataset for knowledge-enhanced urban spatiotemporal prediction. 
More recently, KGs are used in retrieval to generate inputs for LLMs, giving rise to a series of studies like GraphRAG~\cite{edge2024local}. GraphRAG leverages knowledge graphs to enhance Retrieval-Augmented Generation (RAG). For instance, KG2RAG~\cite{zhu2025knowledge} utilizes graph structures for precise context retrieval and multi-hop reasoning to extract connected facts from textual corpora. However, these methods mainly focus on developing RAG techniques for textual data. The integration of knowledge graphs with fine-grained numerical data, often facing granularity discrepancy, is still unexplored.

\textbf{Discussion.} In summary, existing SST prediction methods struggle to integrate critical oceanographic knowledge, a limitation that persists even with powerful LLM-based methods, which are primarily used as generic feature encoders. 
While KGs offer an effective way to represent domain knowledge, current KG-based methods are designed for textual data, facing a significant ``granularity discrepancy" challenge when applied to fine-grained numerical SST data. Therefore, the core contribution of our work is a novel framework that enables LLMs to leverage structured ocean knowledge from a KG to achieve more accurate SST prediction.

\section{Preliminary }

\subsection{Problem Definition}
We represent the historical SST time series data as a matrix $ \mathcal{X} \in \mathbb{R}^{N \times T} $, where $N$ is the number of ocean regions and $T$ is the number of time steps. Each row of this matrix, denoted as $ X_i = (x_{i,1}, x_{i,2}, \dots, x_{i,T}) $, is a vector representing the sequence of temperature values for region $ i $.
The global SST prediction problem is formulated as seeking a function \bm{$\mathcal{F}$} to predict the SST data in upcoming \bm{$\tau$} steps based on the historical data of the last $T$ time points, i.e.,
\begin{equation}
  \mathcal{\hat{Y}} = \bm {\mathcal{F}_{\theta}} \{\mathcal{{X}}\}
 \end{equation}
where \bm{$\theta$} denotes all the learnable parameters in the prediction model, and $\mathcal{\hat{Y}}$ denotes the prediction results. 

Meanwhile, to integrate domain knowledge into prediction, we combine the ocean knowledge graph, denoted as $\mathcal{OKG}$, and numerical data for prediction. Thus, the SST prediction problem is rewritten as follows.
\begin{equation}
    \mathcal{\hat{Y}} = \bm {\mathcal{F}_{\theta}} \{\mathcal{{X}};\bm{\mathcal{OKG}}\}
 \end{equation}

\begin{figure}[]
  \centering
\includegraphics[width=0.5\textwidth]{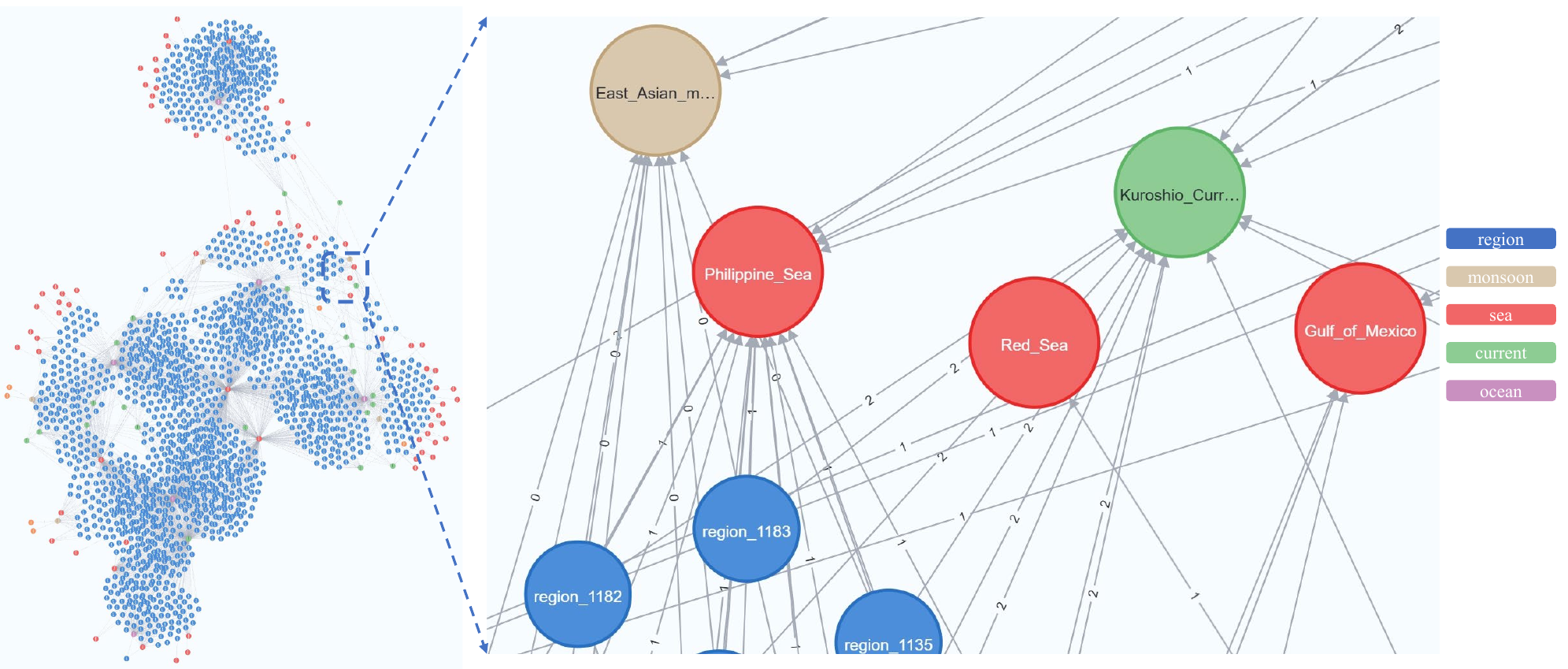}
  \caption{Overview of OKG, where the left panel displays the global topology of the entire graph, and the right panel depicts a specific subgraph.}
    \label{fig:kgvisual}
\end{figure}

\subsection{Construction of Ocean Knowledge Graph }
A comprehensive understanding of the ocean is critical for SST prediction. To integrate disparate ocean knowledge into a structured and queryable format, we construct the Ocean Knowledge Graph (OKG)- a knowledge graph centered on fundamental oceanic concepts.  
We begin by developing the OKG ontology, where the most granular entity type is the Region, representing a specific spatial unit for prediction. Each Region is “located\_in” a larger Sea (e.g., the Arabian Sea), which is “part\_of” a major Ocean (e.g., the Indian Ocean). The ontology also includes influential phenomena such as Ocean Currents and Monsoon Climates. Key relationships connecting these entities include ``located\_in, part\_of, influenced\_by, and adjacent\_to". 
However, directly associating fine-grained regions with broad phenomena like ocean currents and monsoon climates based solely on coordinates~(latitude and longitude) is challenging. To address this, we employ a two-step mapping process. First, we leverage external knowledge sources, such as Wikipedia and the National Oceanic and Atmospheric Administration~(NOAA), to establish the approximate geographic boundaries of these entities. Second, we map each region to the corresponding entities based on its coordinates within these established boundaries. The open-source data and more detailed construction report are available in the online repository. 


\begin{table}[htbp] 
\caption{The statistical information of OKG.} 
\centering 
\begin{tabular}{lc}
\hline
\textbf{Entity} & \textbf{Count} \\
\hline
Regions & 1,715 \\
Currents & 22 \\
Monsoons & 5 \\
Oceans & 6 \\
Sea Areas & 81 \\
\hline
\end{tabular}
\label{tab:kg_info} 
\end{table}

Fig.~\ref{fig:kgvisual} illustrates the visualization of our constructed OKG, and Table~\ref{tab:kg_info} shows its statistical information. The graph is composed of 1,829 distinct entities across five categories. These entities are interconnected by 4,602 triples, which form the factual basis of OKG. 



  


\begin{figure*}[]
  \centering
\includegraphics[width=\textwidth]{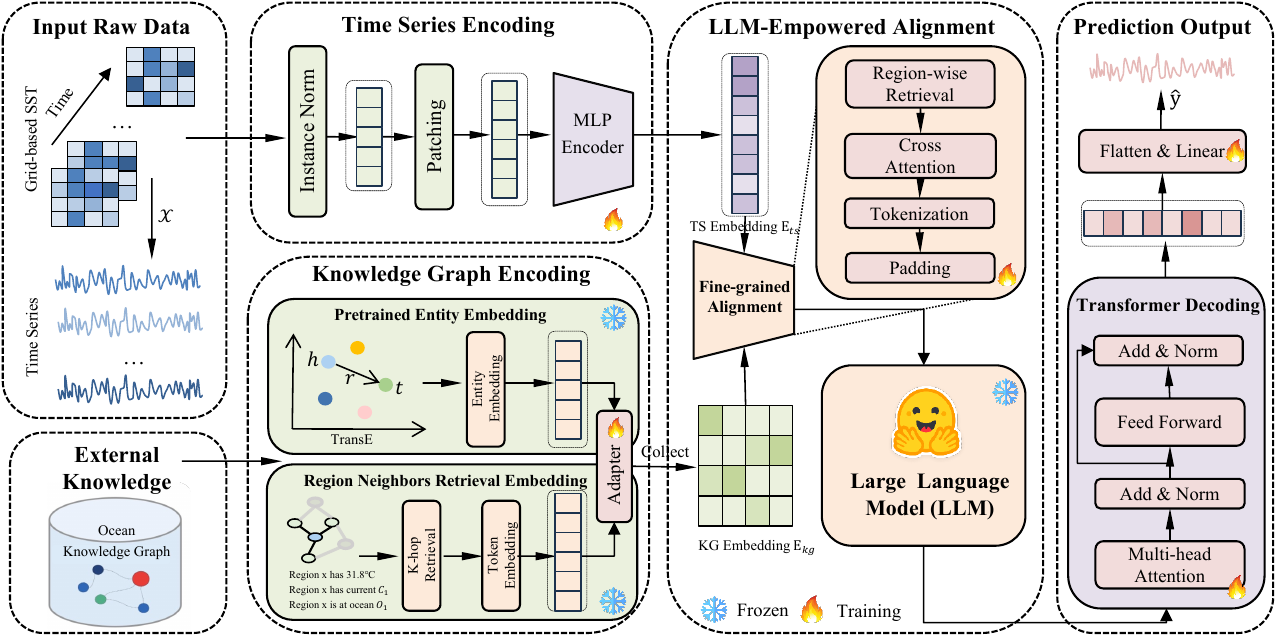}
  \caption{The proposed prediction framework, OKG-LLM, is designed for time series forecasting by unifying observational data with external ocean knowledge. First, time series and ocean knowledge graph data are fed into the Time Series Encoding and Knowledge Graph Encoding modules, respectively, to learn their feature embeddings. Next, the LLM-empowered alignment module utilizes a region-wise retrieval and cross-attention mechanism for alignment, and then introduces a frozen LLM to fuse these multi-modal embeddings. Finally, the Prediction Output module employs a trainable Transformer Decoder to process the aligned representation to generate the final prediction results.}
    \label{model}
\end{figure*}

\section{Method}

\subsection{Framework of OKG-LLM}
Fig.~\ref{model} presents the framework of OKG-LLM, which integrates structured ocean knowledge with fine-grained SST observational data to achieve accurate SST prediction by
leveraging the powerful representation and reasoning capabilities of LLMs.
OKG-LLM consists of four modules. First, the raw SST data are processed by the Time-series Encoding module, which generates temporal feature embeddings. Simultaneously, the Knowledge Graph Encoding module captures and integrates both structural and semantic knowledge of the OKG to produce a unified knowledge-enhanced embedding. Next, the LLM-Empowered Alignment module employs cross-attention to enable fine-grained fusion of the temporal and knowledge embeddings. The fused representation is then fed into a pre-trained LLM to learn high-dimensional patterns.  Finally, a trainable transformer decoder processes the LLM's output to model the spatio-temporal dependencies of SST changes, and the resulting output is passed through a linear layer to produce the final prediction.


\subsection{Time Series Encoding Module}

Given the input SST data $\mathbf{\mathcal{X}} \in \mathbb{R}^{N \times T} $,
the first step is to individually normalize the time series ${X}_i$ for each of the N regions.
We address the distribution shift in the time-series data using Reversible Instance Normalization (RevIN) which standardizes each input sequence to have a zero mean and unit variance~\cite{kim2021reversible}. Subsequently, the normalized time series is segmented into patches to aggregate local features and improve computational efficiency.
Let the patch length be $L_p$ and the stride be $S$. The resulting sequence of patches, ${X}^{(i)}_p \in \mathbb{R}^{P \times L_p}$, has a total number of patches $P = \lfloor(T - L_p)/S\rfloor + 2$. 
We use a {Multi-Layer Perceptron (MLP)} to project these patches into a $d_m$-dimensional embedding space, yielding the final representation ${e}^{(i)}_{ts} \in \mathbb{R}^{P \times d_m}$ for each region, i.e.,
\begin{equation}
{e}^{(i)}_{ts} = \text{MLP}({X}^{(i)}_{p}) = \sigma({X}^{(i)}_{p} {W}_1 + {b}_1){W}_2 + {b}_2
\end{equation}
where ${X}^{(i)}_p$ is the input patch matrix, ${W}_1$ and ${W}_2$ are weight matrices, ${b}_1$ and ${b}_2$ are bias vectors, and $\sigma$ denotes a non-linear ReLU activation function. Finally, we collect the time-series embedding for all regions as ${{E}}_{ts} \in \mathbb{R}^{N \times P \times d_m}$.

\subsection{Ocean Knowledge Graph Encoding Module}

The OKG is denoted as $\mathcal{G} = (\mathcal{E}, \mathcal{R}, \mathcal{T}, \mathcal{D})$ where $\mathcal{E}$ and $\mathcal{R}$ are the entity set and relation set, respectively; $\mathcal{T} = \{(h, r, t) \mid h, t \in \mathcal{E}, r \in \mathcal{R}\}$ is the triple set, and $\mathcal{D}$ is the description set of entities and relations. We denote $\mathcal{D}(h,r,t)$ as the textual description of triple $(h,r,t) \in \mathcal{T} $. 

To augment the time-series data with the ocean knowledge graph, we design a Knowledge Graph Encoding module that distills symbolic information from the OKG into low-dimensional, continuous vector representations. This process enriches the raw numerical data with both structural and semantic knowledge about the geographic entities, which is crucial for a nuanced understanding of oceanic phenomena. The module operates through two parallel branches: \textbf{Pretrained entity embedding} and \textbf{Region neighbors retrieval embedding}.

\textbf{Pretrained entity embedding for regional inter-correlations.}
The primary objective of this sub-module is to capture the regional inter-correlations in the OKG.
We employ TransE, a translational knowledge graph embedding model, to project entities and relations into a shared low-dimensional space. For a given triple $(h, r, t)$, the embedding of the tail entity ${t}$ should be a translation of the head entity's embedding ${h}$ by the relation's vector ${r}$, i.e., ${h} + {r} \approx {t}$.

The TransE model is pre-trained on OKG and optimized using a margin-based ranking loss function $\mathcal{L}$, i.e., 
\begin{equation}
\mathcal{L} = \sum_{(h,r,t) \in \mathcal{T}^+} \sum_{(h',r,t') \in \mathcal{T}^-_{(h,r,t)}} \left[ \gamma + f(h,r,t) - f(h',r,t') \right]_+
\end{equation}
where $[x]_+ = \max(0, x)$, $\gamma$ is the margin hyperparameter, $\mathcal{T}^+$ represents the set of positive triples (observed facts in the knowledge graph), and $\mathcal{T}^-_{(h,r,t)}$ is the set of negative triples generated by replacing either the head $h$ or tail $t$ of a positive triple $(h,r,t)$ with a randomly selected entity.
The scoring function $f(h,r,t) = \|{h} + {r} - {t}\|_p$ (typically with $p=1$ or $p=2$) measures the plausibility of a triple. After pre-training, each entity $e \in \mathcal{E}$ (e.g., a specific grid point, an ocean, or a current) is represented by a global, structure-aware embedding ${e}_{\text{struct}} \in \mathbb{R}^{d}$ 
Such embeddings for entities are typically \text{frozen} during the training for SST prediction model. This strategy not only preserves the learned general-purpose structural knowledge, but also improves training efficiency by reducing the number of trainable parameters.

\begin{figure}[]
  \centering
\includegraphics[width=0.5\textwidth]{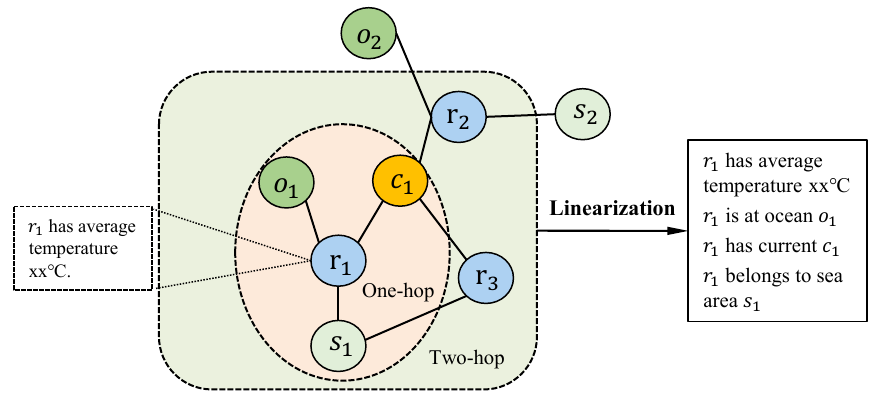}
  \caption{Region Neighbors Retrieval Embedding.}
    \label{fig:khop}
\end{figure}

\textbf{Region neighbors retrieval embedding for regional semantic knowledge.}
While the pretrained entity embeddings capture regional intercorrelations, they may underrepresent the specific semantic attributes of an entity. This sub-module addresses this limitation by verbalizing the entity's local graph structure into a natural language prompt, which is then is then processed by an LLM to obtain a fine-grained semantic representation. The process involves two steps.
\begin{enumerate}
    \item \textbf{k-hop retrieval:} For a given target entity $e$, we first perform a k-hop retrieval to collect its local neighborhood graph $\mathcal{N}_K(e)$.  The triples in this neighborhood are linearization into a textual document $\text{Text}(e)$. The region entity's description is derived from the average sea surface temperature of that region over a specific period. Meanwhile, we concatenate the entity’s description $\mathcal{D}(e)$, i.e.,
    \begin{equation}
    \text{Text}(e) = \text{Concat}\left(\mathcal{D}(e); \mathcal{D}(h,r,t) \mid (h,r,t) \in \mathcal{N}_K(e) \right)
    \end{equation}
    For example, as shown in Figure~\ref{fig:khop}, $\text{Text}(r_1 )$ is the concatenation of \text{($r_1$, has temperature, xx$^{\circ}$C)}, \text{($r_1$, is at, ocean $o_1$)}, \text{($r_1$, belongs to, sea area $s_1$)} and \text{($r_1$, has current, $c_1 $)}
    In practice, we strike a balance between computational efficiency and performance by leveraging 1-hop neighbors, which effectively capture the entity's localized semantic features.

    \item \textbf{Token embedding:} The retrieved text $\text{Text}(e)$ is then fed into a LLM to generate a dense \text{token embedding} $\mathbf{e}_{\text{text}} \in \mathbb{R}^{l\times d}$ that serves as a rich semantic summary of the entity's specific properties and direct relationships, i.e.,
    \begin{equation}
    {e}_{\text{text}} = \text{LLM}_{token}(\text{Text}(e))
    \end{equation}  
\end{enumerate}

\textbf{Knowledge fusion via an adapter.}
Finally, the two distinct representations, i.e., the global, structure-aware entity embedding $\mathbf{e}_{\text{struct}}$ and the local, description-rich token embedding $\mathbf{e}_{\text{text}}$, are integrated. Inspired by KOPA~\cite{zhang2024making}, they are fed into a lightweight, trainable \text{adapter}, implemented as a small MLP, to learn the optimal fusion strategy. The adapter dynamically weighs and combines the two heterogeneous information sources to produce a single, unified, knowledge-enhanced vector $\mathbf{e}_{\text{kg}} \in \mathbb{R}^{l\times d}$ that serves as the final output of ocean knowledge graph encoding module, i.e.,
\begin{equation}
{{e}}_{\text{kg}} = \text{Adapter} \left( [{e}_{\text{struct}} ; {e}_{\text{text}}] \right) = \sigma \left( {W} [{e}_{\text{struct}} ; {e}_{\text{text}}] + {b} \right)
\end{equation}
where $[\cdot ; \cdot]$ denotes vector concatenation, ${W}$ and ${b}$ are the trainable weight matrix and bias vector, respectively, of the MLP, and $\sigma$ is the non-linear activation function ReLU.
Finally, we collect the embeddings for all ocean regions in OKG $\mathcal{kG}$ to yield a comprehensive feature set $E_{\text{kg}} \in \mathbb{R}^{N\times l \times d}$ which will be used to augment the SST series features.

\subsection{Fine-grained Alignment with LLM Module} 


To accurately predict SST, it is crucial to effectively fuse its distinct temporal embeddings with contextual ocean knowledge embeddings. To this end, we design an LLM-empowered alignment module. First, this module aligns and merges the knowledge-based embeddings and temporal embeddings to create unified, context-aware representations for each ocean region. Then, it leverages a pre-trained LLM to capture the SST patterns of each region within a high-dimensional space. 


Specifically, for each region $i$, we first retrieve its unique region entity embedding $e_{\text{region}}^{i}$ and prepend it to the corresponding temporal embedding $e_{\text{ts}}^{i}$. This forms a context-aware query vector:
\begin{equation}
 e_{\text{query}}^{i} = \text{LayerNorm} (\text{concat}[e_{\text{region}}^{i}, e_{\text{ts}}^{i}])
\end{equation}
This query then attends to the broader knowledge graph embeddings $E_{kg}$, which serve as the 'Key' and 'Value'. 
\begin{equation}
E_{\text{aligned}} \\ = softmax\left(\frac{(E_{\text{query}} W_Q)(E_{\text{kg}} W_K)^T}{\sqrt{d_k}}\right)(E_{\text{kg}} W_V)
\end{equation}
where $E_{\text{aligned}}$ is the learned embedding, $W_Q, W_K, W_V$ are the learnable weight matrices, and $d_k$ is the vector dimensionality used for scaling. 


To learn complex patterns from the aligned embeddings, we employ a pre-trained Large Language Model (LLM). We chose decoder-only models as our backbone due to their autoregressive nature, which uses causal masking to preserve the temporal order of the data. For feature extraction, we use a pre-trained LLM (such as GPT-2, Llama,  OceanGPT, or DeepSeek) with all its parameters frozen. Instead of using the model's native tokenizer, we treat our aligned embeddings, $E_{\text{aligned}}$, as direct inputs to the transformer layers. These input sequences are first padded to a uniform length, allowing the model to process them and generate high-level feature representations.
 

\subsection{Prediction Output Projection Module}

After obtaining the high-level feature representations from the LLM, a trainable transformer decoder uses multi-head self-attention to refine the representations for downstream SST prediction performance, i.e.,
 \begin{gather}
  Attention{(Q,K,V)} = softmax(\frac{QK^T}{\sqrt{d_k}}) \\
  MultiHead(Q,K,V) = Concat(head_1, \cdots ,head_h)  \\
  head_i = Attention(QW^q_i,KW^k_i,VW^v_i) \notag
 \end{gather}
where matrices $Q, K, V$ are all derived from the LLM's output features, $W$ represents the learnable parameters of the transformer block, and $MultiHead()$ concatenates the information from different representation subspaces $(head_1, \cdots ,head_h)$ to learn diverse temporal patterns.

Finally, the linear layer bridges the gap between the representation space and the prediction space. It takes the high-dimensional, contextualized vectors from the decoder and projects them into the final task-specific output form.

For the training objective, we use the L1 loss function to minimize the difference between the model's predictions and the ground truth values, i.e.,
 \begin{equation}
      {\mathcal{L}_\text{pred}} =  {\mathcal{L}({\mathcal{F}_\theta})} =\frac{1}{\tau} \sum_{i=0}^\tau{\left|{{\hat{{\mathcal{Y}}}_{i}-{{\mathcal{Y}_i}}}}\right|}
 \label{loss1}
  \end{equation}
where {$\theta$} denotes all the learnable parameters, and $\tau$ is the total time steps need to be predicted.

\section{Experiments}
We conduct experiments on real-world global datasets to demonstrate the superiority of the proposed method against multiple strong baseline methods.
We also evaluate the impacts of the hyper-parameters and the effectiveness of model components.
Moreover, we present visualizations to showcase the outcomes of our method as well as the baseline approaches, providing clear and concise comparisons. 

\subsection{Dataset Details}
In this study, to evaluate the performance of OKG-LLM, we utilize a real-world and commonly used SST observation dataset from NOAA\footnote{{Dataset available from: https://psl.noaa.gov}}. This dataset provides weekly global SST data from 1991 to the present, and has a spatial resolution of 5° x 5°, encompassing approximately 1,800 oceanic grid cells.

\subsection{Baselines}

We compare our method, OKG-LLM, with nine state-of-the-art models. The baselines consist of two groups, i.e., five typical forecasting models, including Informer, FEDformer, DLinear, FreTS, and TimeMixer+, and four advanced LLM-based models including GPT4TS, UniTime, TimeLLM, and TimesFM.

\begin{itemize}
\item {\bfseries Informer}~(Zhou et al. 2021~\cite{zhou2021informer}): An efficient Transformer-based model for long sequence time-series forecasting, featuring a ProbSparse self-attention mechanism and a self-attention distilling operation to reduce complexity and manage long input sequences.
\item {\bfseries FEDformer}~(Zhou et al. 2022~\cite{zhou2022fedformer}): A transformer-based model that combines seasonal-trend decomposition with frequency domain learning, employing Fourier Enhanced Block and Wavelet Enhanced Block for capturing global and local patterns, respectively.
\item {\bfseries DLinear}~(Zeng et al. 2023~\cite{zeng2023transformers}): A simple yet effective linear model for time series forecasting, which decomposes the time series into trend and seasonal components, and applies separate linear layers to them, challenging the necessity of using complex transformer-based models for some forecasting tasks.
\item {\bfseries FreTS}~(Yi et al. 2023~\cite{yi2023frequency}): A model that disentangles time series into trend and seasonal/residual components, and applies a frequency-domain analysis to the seasonal/residual component and a simple polynomial fitting for the trend component, aiming for efficient and interpretable forecasting.
\item {\bfseries GPT4TS}~(Zhou et al. 2024~\cite{zhou2023one}): A time series forecasting model based on pre-trained Generative Pre-trained Transformer (GPT) models, which tokenizes time series data and fine-tunes the large language model for forecasting tasks, leveraging the strong sequence modeling capabilities of LLMs. 
\item {\bfseries UniTime}~(Liu et al. 2024~\cite{liu2024unitime}): A unified time series model based on a Transformer decoder architecture, capable of handling diverse time series tasks including forecasting, imputation, and classification through multi-task learning on large-scale datasets.
\item {\bfseries TimeLLM}~(Jin et al. 2024~\cite{jin2023time}): A framework that reprograms LLMs for time series forecasting by aligning time series data with the input format of LLMs using text prototypes and then fine-tuning LLMs for the forecasting task. 
\item {\bfseries TimeXer++}~(Wang et al. 2025~\cite{wang2024timemixer}): A general time series pattern mining architecture using specialized techniques to model diverse temporal patterns for universal predictive analysis.
\item {\bfseries TimeFM}~(Google et al. 2025~\cite{das2024decoder}): A decoder-only Transformer foundation model from Google, pre-trained on 100 billion time series data points. It directly applies an LLM architecture to time series forecasting, differing from approaches that reprogram general-purpose LLMs using text prototypes of time series.
\end{itemize}

\begin{table*}
  \caption{The overall results of OKG-LLM and baseline models on predicting global SST with three different prediction lengths. The results show that OKG-LLM consistently outperforms both typical models and LLM-based models.}
  \label{result-table}
  \centering
  \normalsize
  \renewcommand{\arraystretch}{1.2}
  \begin{tabular}{c l cc cc cc}
    \hline
    & \multirow{2}{*}{Methods} & \multicolumn{2}{c}{$\tau = 8$} & \multicolumn{2}{c}{$\tau = 16$} & \multicolumn{2}{c}{$\tau = 32$} \\
    \cline{3-8} 
    & & MSE & MAE & MSE & MAE & MSE & MAE \\
    \hline
    \multirow{5}{*}{\rotatebox{90}{Typical}} 
    & Informer (2021) & 0.2762 & 0.3751 & 0.2802 & 0.3772 & 0.2905 & 0.3804 \\
    & FEDformer (2022) & 0.2046 & 0.3233 & 0.2092 & 0.3283 & 0.217 & 0.334 \\
    & DLinear (2023) & 0.1455 & 0.2291 & 0.1622 & 0.2956 & 0.2034 & 0.3294 \\
    & FreTS (2024) & 0.1162 & 0.2605 & 0.1883 & 0.2712 & 0.2354 & 0.3061 \\
    & TimeMixer+ (2025) & 0.1132 & 0.2495 & 0.1592 & 0.2825 & 0.1981 & 0.2952 \\
    \hline
    \multirow{4}{*}{\rotatebox{90}{LLM-based}} 
    & GPT4TS (2023) & 0.1254 & 0.2514 & 0.1592 & 0.2767 & 0.1796 & 0.2967 \\
    & UniTime (2024) & 0.1194 & 0.2413 & 0.1514 & 0.2697 & 0.1772 & 0.2916 \\
    & TimeLLM (2024) & \underline{0.1063} & \underline{0.2210} & \underline{0.1402} & \underline{0.2526} & 0.1700 & 0.2803 \\
    & TimesFM (2025) & 0.1091 & 0.2351 & 0.1526 & 0.2872 & \underline{0.1681} & \underline{0.2745} \\
    \hline

    & \bf{OKG-LLM (Ours)} & \bf{0.0982} & \bf{0.2071} & \bf{0.1326} & \bf{0.2414} & \bf{0.1626} & \bf{0.2680} \\
    \hline
    \label{tab:performance-comparison}
  \end{tabular}
\end{table*}

\subsection{Experimental Setups}
We briefly introduce the prediction scheme, experimental environment, parameter setting, and evaluation metrics.

{\bfseries Prediction scheme.} {All methods predict the SST for future $\tau$ time steps based on the historical observations in past $T$ time steps. For example, given the setting with $T=8$ and $\tau=8$, the model utilizes the historical SST values of the past 8 weeks to predict the SST of next 8 weeks.
In the experiments, we cover three prediction lengths, i,e, 8, 16 and 32, following the previous studies~\cite{zhou2021informer} to provide a full comparison between our model and baseline methods.}

{\bfseries Experimental environment.} All the models are implemented in Python with Pytorch 1.13.1, where the source code in the original papers is used for the baseline methods.
We run all the models on a server with four NVIDIA 4090 GPUs, and optimize them by the Adam optimizer with a maximum of 50 epochs and an early stopping strategy to avoid overfittings.

{\bfseries Parameter setting.}  We repeat each experiment ten times, and the best parameters for all prediction models are chosen through a careful parameter-tuning process. 
The batch size is set to 64, 16, and 16 for the experiments with prediction lengths of 8, 16, and 32, respectively.
The original learning rate is set to 0.0001 and halved every four epochs. 

{\bfseries Evaluation metrics}. We use two widely used metrics, i.e., Mean Absolute Error (MAE) and Mean Square Error (MSE), to measure the performance of prediction models i.e..
\begin{equation}
  \begin{split}
   \textrm{MAE}(\mathcal{\hat{Y}}, \mathcal{{Y}}) &= \frac{1}{ \tau} \sum_{i=1}^{\tau} \left| Y_i - \hat{Y}_i \right| \\
   \textrm{MSE}(\mathcal{\hat{Y}}, \mathcal{{Y}}) & = \frac{1}{\tau} \sum_{i=1}^{\tau} (Y_i - \hat{Y}_i)^2
   \end{split}
   \label{t}
\end{equation}
where ${\mathcal{Y}} = \{Y_{:1}, Y_{:2},\cdots, Y_{:\tau}\}$ denotes the ground truth, ${\mathcal{\hat{Y}}} = \{\hat{Y_{:1}}, \hat{Y_{:2}},\cdots, \hat{Y_{:\tau}\}}$ represents the predicted values, and $\tau$ denotes the  time steps to be predicted.
In our experiments, $\tau$ is set to 8, 16, and 32, respectively.

\subsection{Experimental Results}

Table~\ref{tab:performance-comparison} summarizes the results of 9 baseline models and our proposed OKG-LLM on global SST prediction tasks with three different prediction lengths.
Based on the results, we derive the following observations: (1) OKG-LLM consistently outperforms all baseline methods across all metrics and prediction lengths. Compared to TimeLLM, the second-best model, OKG-LLM, achieves an obvious improvement of 7.5\% to 15.1\% in terms of MSE across all prediction lengths. This highlights the effectiveness of OKG-LLM in modeling complex temporal dependencies in SST data.
(2) LLM-based approaches (e.g., GPT4TS and TimeLLM) demonstrate better performance than traditional transformer-based methods (e.g., Informer and FEDformer). However, OKG-LLM surpasses even the best-performing LLM-based approaches as it benefits from a hybrid design that combines domain-specific priors with advanced neural architectures.
(3) As the prediction length increases, OKG-LLM maintains its robustness. While most baseline methods show a noticeable decline in performance with longer forecasting sequences, OKG-LLM consistently achieves superior results. This demonstrates its scalability and effectiveness in handling long-range SST prediction tasks.

Table~\ref{tab:efficiency-comparison} highlights that OKG-LLM achieves a well-balanced trade-off between performance and efficiency. By comparing total parameters, memory usage, and per-batch processing time across LLM-based methods, we observe that OKG-LLM delivers superior prediction accuracy with manageable computational costs. It significantly reduces overhead compared to heavy models like TimeLLM, while slightly exceeding GPT4TS in size yet far surpassing it in accuracy. This balance makes OKG-LLM both practical and highly effective for real-world applications.

In summary, the above results clearly indicate that OKG-LLM sets a new benchmark for global SST prediction, outperforming both traditional time-series models and state-of-the-art LLM-based approaches. It achieves excellent prediction performance by fully integrating the multi-scale ocean knowledge into LLM and aligning the knowledge with fine-grained numerical SST data for prediction.

\begin{table}[]
  \centering
  \caption{Efficiency analysis of OKG-LLM and LLM-based baselines.}
    \renewcommand{\arraystretch}{1.1} 
    \normalsize
  \begin{tabular}{cccc}
    \hline
    Methods &  \#Parameters                & \#Memory & \#Speed(Second) \\
    \hline
   GPT4TS      & \textbf{5.52}             & \textbf{1,021}    & \textbf{0.18}  \\
   TimeLLM     & 148.26            & 29,182    & 3.18    \\
    UniTime      & 50.54            & 5,728    & 0.88    \\
    TimesFM      & \underline{24.52}       &  2,245    & 0.36   \\    
    \textbf{OKG-LLM}      & 40.15       &  \underline{1,345}    & \underline{0.21}   \\    \hline  
  \end{tabular}
  \label{tab:efficiency-comparison}
\end{table}

\subsection{Ablation Study}

To evaluate the effectiveness of the key components in OKG-LLM, we conducted a detailed ablation study. Specifically, we compared the full OKG-LLM model against its three variants, each with a specific component removed:
\begin{itemize}
\item {}{\bfseries w/o Time Series Encoding}: deleting the MLP layer in the time series encoding module, which captures the temporal correlations in the SST data. 
\item {}{\bfseries w/o Knowledge Graph Encoding}: removing the knowledge data input and deleting the knowledge graph encoding module to predict only with time series data.
\item {}{\bfseries w/o Fine-grained Alignment}: utilize the concat module instead of the region-wise retrieval and cross-attention module for the combination of ocean knowledge embedding and time series embedding.
\end{itemize}

\begin{figure}[]
  \centering
\includegraphics[width=0.5\textwidth]{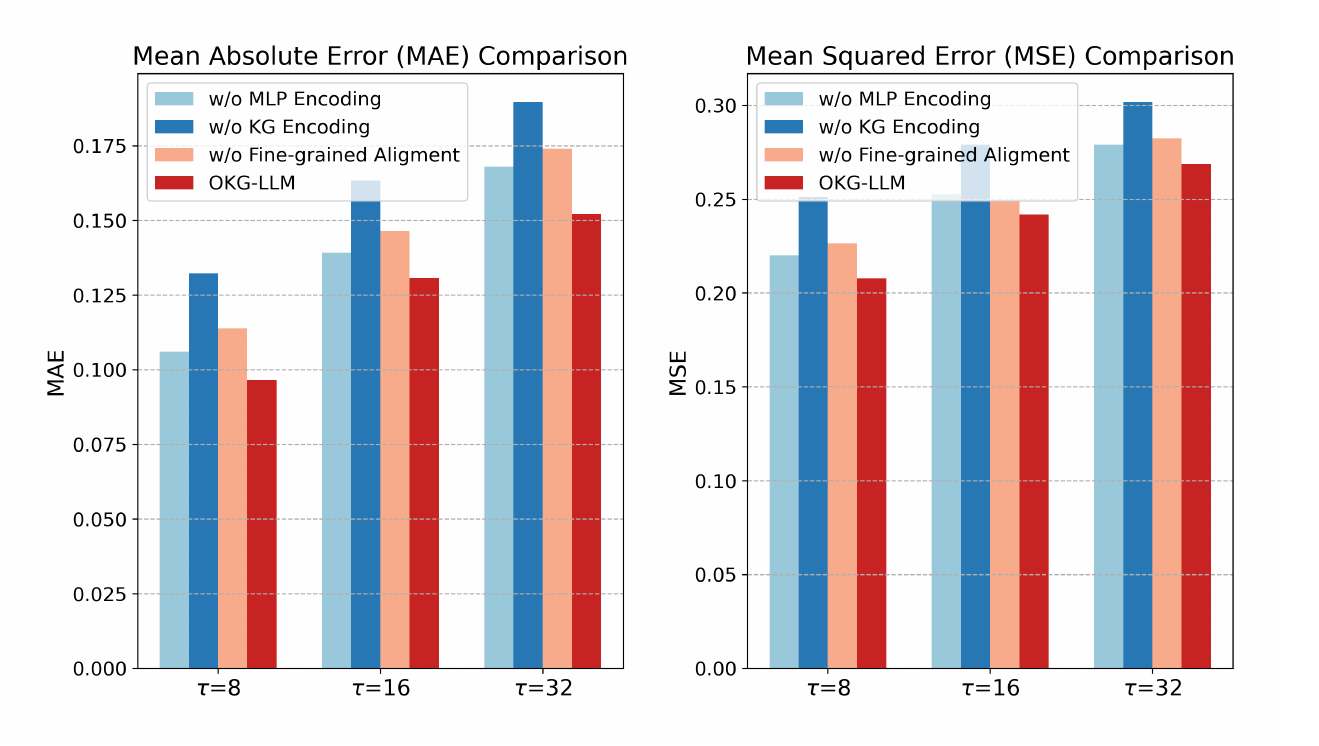}
  \caption{The ablation study on different variants of OKG-LLM.}
    \label{abl}
\end{figure}

The performance of these variants is benchmarked against the complete OKG-LLM model using MAE and MSE metrics across three prediction horizons.
As shown in Fig.~\ref{abl}, the full OKG-LLM model consistently outperforms all ablated versions across all tested configurations, achieving the lowest MAE and MSE. The removal of the KG Encoding results in the most significant performance degradation, indicating that this component is fundamental to the model's predictive capability. Similarly, removing the MLP Encoding also leads to a substantial increase in prediction error, underscoring the importance of integrating knowledge graph information for accurate forecasting.
Furthermore, the model w/o Fine-grained Alignment also shows a noticeable drop in performance compared to the full model. This suggests that the alignment mechanism provides a valuable contribution to refining the model's final predictions. The consistent trend across both MAE and MSE metrics confirms the positive impact of each component.

The ablation study demonstrates that the MLP Encoding, KG Encoding, and Fine-grained Alignment modules are all essential and integral components, each contributing to the superior performance of the OKG-LLM model.

\begin{table}[]
  \centering
  \caption{Performance comparison of different LLM backbones.}
    \renewcommand{\arraystretch}{1.2} 
    \normalsize
  \begin{tabular}{cccc}
    \hline
    LLM backbones & Methods                & Avg MSE & Avg MAE \\
    \hline
    \multirow{4}{*}{GPT-2}      & GPT4TS             & 0.154    & 0.274    \\
   ~      & UniTime            & 0.149    & 0.267    \\
    ~       & TimeLLM            & 0.145    & 0.265    \\
     ~       & OKG-LLM       & \textbf{0.131}    & \textbf{0.239}    \\     \hline
    \multirow{2}{*}{Llama2-7b}     & TimeLLM             & \textbf{0.136}    & 0.251    \\
    ~  & OKG-LLM      & 0.138    & \textbf{0.250}    \\     \hline
    \multirow{2}{*}{Deepseek-R1}     & TimeLLM             & 0.143    & 0.268    \\
    ~     & OKG-LLM      & \textbf{0.141}    & \textbf{0.258}    \\
    
    \hline
    \multirow{2}{*}{OceanGPT}     & TimeLLM             & 0.144   & 0.260    \\
    ~     & OKG-LLM      & \textbf{0.137}    & \textbf{0.253}    \\

    \hline
  \end{tabular}
  \label{tab:llm_comparison_avg}
\end{table}

\begin{figure*}[]
  \centering
\includegraphics[width=\textwidth]{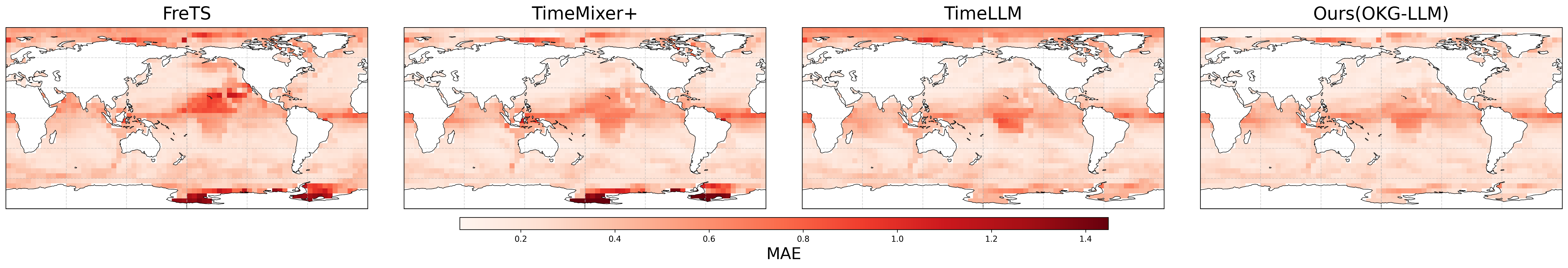}
  \caption{Spatial visual comparison of MAE on different prediction models, i.e., FreTS, TimeMixer+, TimeLLM, and OKG-LLM, where the color indicates the MAE and lighter shades represent lower error.}
    \label{vis}
\end{figure*}

\begin{figure*}[]
  \centering
\includegraphics[width=\textwidth]{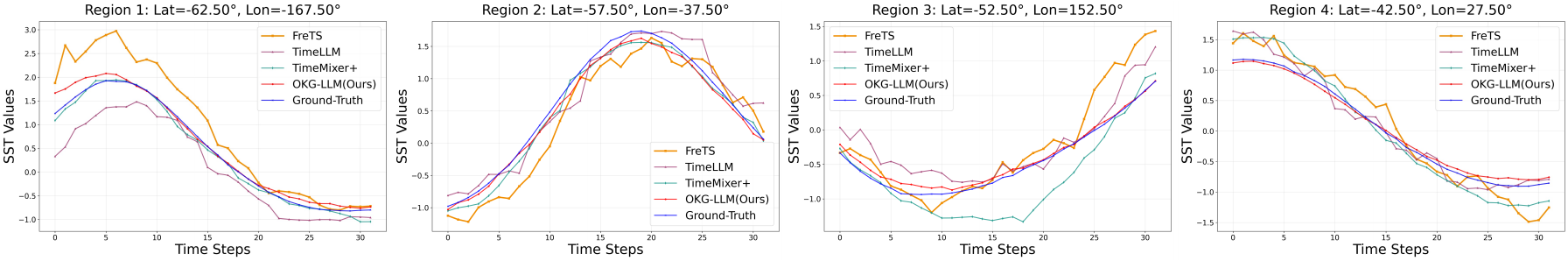}
  \caption{Temporal visual comparison of ground truth and the predicted values from different prediction models, i.e., FreTS, TimeMixer+, TimeLLM, and OKG-LLM, in four randomly selected oceanic regions (the Southern Ocean, the Scotia Sea, the Gulf of Alaska, and the mid-south Atlantic Ocean), where the predictions of our OKG-LLM model are the most similar to the ground truth.}
    \label{vispoint}
\end{figure*}

\subsection{LLM Backbone Comparison}

To validate the generalizability and robustness of our proposed OKG-LLM framework, we integrated it with a diverse set of representative LLMs, including three general models (GPT-2, Llama2-7b, and Deepseek-R1) and one domain-specific model, i.e., OceanGPT~\cite{bi2024oceangpt}, which targets ocean-related NLP tasks. The results in Table~\ref{tab:llm_comparison_avg} underscore the adaptability of our method.

The empirical results reveal that OKG-LLM consistently enhances the prediction performance regardless of the underlying LLM. When paired with GPT-2, our framework achieves the best performance, significantly outperforming the baselines. This performance advantage is maintained with Deepseek-R1 as the backbone, where OKG-LLM again demonstrates superior results compared to TimeLLM in both MSE and MAE metrics.
With the Llama2-7b backbone, TimeLLM records a slightly lower MSE, but our OKG-LLM framework still secures a better MAE and delivers competitive, well-balanced performance. 
When utilizing the domain-specific OceanGPT as the backbone, our method continues to outperform the baselines, confirming that integrating our OKG module provides a greater performance benefit than fine-tuning the LLM on domain-specific knowledge for SST prediction.

Collectively, these findings affirm that OKG-LLM is not tailored to a specific LLM model, but functions as a versatile and effective framework, capable of delivering state-of-the-art or highly competitive results across a diverse range of LLMs.

\subsection{Results Visualization}

Fig.~\ref{vis} presents a comparative visualization of the MAE of OKG-LLM against three representative baselines. As illustrated, the LLM-based approaches (TimeLLM and our OKG-LLM) demonstrate a clear superiority over traditional methods like FreTS and TimeMixer+, exhibiting lower overall prediction errors across the globe. In contrast, traditional methods exhibit significant performance degradation in specific regions, as indicated by dark red patches representing high MAE. More importantly, when comparing with LLM-based models, OKG-LLM achieves noticeably lower MAE than TimeLLM. This improvement is particularly pronounced in climate-sensitive areas such as the El Niño-Southern Oscillation (ENSO) region in the equatorial Pacific, underscoring the importance of integrating structured oceanographic knowledge for SST prediction.

To further validate these findings at a granular level, Fig.~\ref{vispoint} visualizes the prediction performance of our model against baselines in four randomly selected regions. The results clearly show that OKG-LLM's predictions (red line) consistently align more closely with the ground-truth data (blue line) than all competing methods.
This superior performance underscores the effectiveness of leveraging knowledge graph OKG to better understand complex regional oceanic patterns for accurate SST forecasting.

\subsection{Embedding Visualization}

\begin{figure*}[]
  \centering
\includegraphics[width=\textwidth]{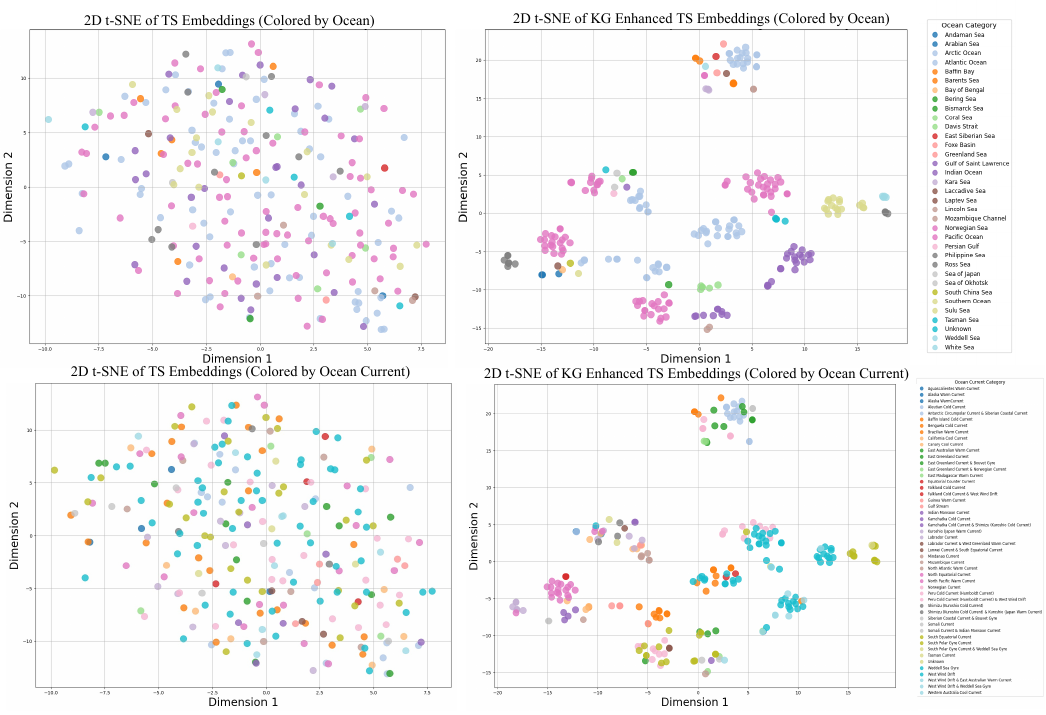}
  \caption{Comparison of embeddings visualized with t-SNE. Our Knowledge Graph~(KG) Enhanced Times series~(TS) Embeddings (right) form distinct clusters for different oceans and currents, while the baseline TS Embeddings (left) do not.}
    \label{tse}
\end{figure*}

As presented in Fig.~\ref{tse}, we visualize the learned embedding spaces using t-SNE to verify whether our framework produces more meaningful and discriminative representations than the baseline that relies solely on time series information.

The left panels of  Fig.~\ref{tse} illustrate the embedding space of a traditional time series model of SST data. The resulting chaotic distribution (no clear separation between different oceans or ocean Currents) suggests that the model fails to learn the semantic similarity between related time series points. These representations lack the necessary structure to distinguish, for example, between data from the Atlantic Ocean versus the Pacific Ocean.
In contrast, the right panels showcase the embedding space generated by our OKG-LLM. The visualizations demonstrate a clear and compelling structure. Data points belonging to the same semantic category are mapped to well-defined clusters in the embedding space characterized by high intra-cluster similarity and low inter-cluster similarity.

This clear difference provides compelling visual proof of our method's efficacy. It confirms that by integrating a knowledge graph, OKG-LLM successfully fuses the numerical temporal representations with rich, structured domain knowledge. This integration results in a semantically meaningful feature space that enhances the SST prediction performance.

\section{Conclusion}
This paper proposes OKG-LLM, a novel framework that advances SST prediction by systematically integrating domain knowledge and observational data.
The key innovation lies in constructing the structured Ocean Knowledge Graph (OKG)—which encodes physical oceanographic principles—and developing an effective fusion mechanism to align this symbolic knowledge with numerical SST data using LLMs. 
Extensive experiments show that OKG-LLM consistently outperforms state-of-the-art baseline models across all evaluation metrics. Ablation studies also confirm that each component, particularly the knowledge graph encoding and fine-grained alignment, is critical to the model's superior performance. This work demonstrates that combining domain knowledge with LLMs is a highly effective and robust approach for complex ocean prediction tasks.
As for future work, it would be interesting to further enhance SST prediction by incorporating various complex environmental knowledge, including air-sea interaction and carbon cycling.

\section*{Acknowledgments}
This work was supported in part by the National Natural Science Foundation of China (No. 62202336, No. 62172300, No. 62372326), Hong Kong Research Grants Council Theme-based Research Scheme (T22-502/18-R), Hong Kong Research Grants Council Research Impact Fund (R5006-23), and the Research Institute for Artificial Intelligence of Things, The Hong Kong Polytechnic University.


%

\bibliographystyle{IEEEtran}
\bibliography{ref}

\end{document}